%% file: root.tex
\documentclass[letterpaper, 10 pt, conference]{ieeeconf}  % Comment this line out if you need a4paper

\IEEEoverridecommandlockouts                              % This command is only needed if 
\input{commands}

\usepackage{graphicx}
\usepackage{afterpage}
\usepackage{stmaryrd}
\usepackage{colortbl}   % pour colorer les cellules
\usepackage{xcolor}     % pour définir les couleurs
\usepackage{float}
\usepackage{balance}
\usepackage{cite}
\usepackage{multirow}

\usepackage[font=small,labelfont=bf]{caption}
\usepackage[font=small,labelfont=bf]{subcaption}
\definecolor{darkgreen}{RGB}{0,100,0} 
\definecolor{darkyellow}{RGB}{204,153,0} 
\definecolor{darkorange}{RGB}{204,85,0}   
\definecolor{darkblue}{RGB}{0,100,200} 
\definecolor{darkred}{RGB}{139,0,0}

\usepackage{amssymb}

\title{\LARGE \bf
Failure Identification in Imitation Learning  \\ via Statistical and Semantic Filtering
}

\author{Quentin Rolland$^{1,2}$, Fabrice Mayran de Chamisso$^{1}$, Jean-Baptiste Mouret$^{2,3}$%
\thanks{$^{1}$Université Paris-Saclay, CEA, List, F-91120, Palaiseau, France} 
\thanks{$^{2}$Inria, CNRS, Université de Lorraine, LORIA, F-54000 Nancy, France}
\thanks{$^{3}$Bleu Robotics, Paris, France}
\thanks{correspondance to: \texttt{quentin.rolland2@cea.fr}.}
\thanks{
This publication was made possible by the use of the CEA List FactoryAI supercomputer, financially supported by the Ile-de-France Regional Council. This work was partly funded by :
 the European Union’s Horizon Europe Research and Innovation program under grant agreement n\textsuperscript{o} 101135708 (JARVIS project), n\textsuperscript{o}101070227 (CONVINCE project) and n\textsuperscript{o} 101070596 (euROBIN project) 
 - and by the France 2030 program through the PEPR O2R projects AS3 (ANR-22-EXOD-007).
}}

\begin{document}

%\thanks{*This work was not supported by any organization}% <-this % stops a space

%\thanks{$^{1}$Université Paris-Saclay, CEA, List, Palaiseau, France \;\;\;\;\;\;\;\;\;\;\;}\\
%\thanks{\; \; \; \; \; \; $^{2}$Inria, CNRS, Université de Lorraine, France}

\addeditor{jeanBaptiste}{JB}{1, 0.4, 0.0}
\addeditor{fabrice}{FAB}{0.6, 0, 1}
\addeditor{quentin}{Q}{0, 0.5, 0.5}

%\showeditstrue
\showeditsfalse

\maketitle

\input{sec/0_abstract}
\input{sec/1_intro}
\input{sec/2_related_work}

\input{sec/3_method}

\input{sec/4_evaluation}
\input{sec/5_conclusion}
\bibliographystyle{IEEEtran}
\bibliography{refs}

\end{document}

%% file: commands.tex
\usepackage[dvipsnames]{xcolor}

\definecolor{lightblue}{RGB}{100, 149, 237}

\usepackage{colortbl}
\usepackage{multirow}
\usepackage[para]{footmisc}
\usepackage{lipsum}
\usepackage{booktabs}

\usepackage{xspace}
\usepackage{hyperref}

\makeatletter
\DeclareRobustCommand\onedot{\futurelet\@let@token\@onedot}
\def\@onedot{\ifx\@let@token.\else.\null\fi\xspace}

\makeatother

\usepackage{amsmath}
\usepackage{amssymb}
\usepackage{dsfont}
\usepackage{etoolbox}
\usepackage{color}
\usepackage{ulem}

\newif\ifshowedits

\newcommand{\addeditor}[3]{%
  \definecolor{#1color}{rgb}{#3}
  \expandafter\newcommand\csname #1\endcsname[1]{%
  \ifshowedits
    {\color{#1color} ##1}%
  \else
    {##1}%
  \fi
  }%
  \expandafter\newcommand\csname #1rmk\endcsname[1]{%
  \ifshowedits
    {\color{#1color} {\bf [#2: ##1]}}
  \fi
  }%
  \expandafter\newcommand\csname #1rpl\endcsname[2]{%
  \ifshowedits
    {\color{#1color} ##1 \sout{##2}}
  \else
    {##1}
  \fi
  }%
}

\newcommand{\createtextvar}[1]{
  \expandafter\newcommand\csname #1\endcsname{%
  {\text{#1}}
}%
}
\newcommand{\mycomment}[1]{}

\newcommand{\vcomment}[1]{}

%% file: sec/0_abstract.tex
\begin{abstract}

Imitation learning (IL) policies in robotics deliver strong performance in controlled settings but remain brittle in real-world deployments: rare events such as hardware faults, defective parts, unexpected human actions, or any state that lies outside the training distribution can lead to failed executions. Vision-based Anomaly Detection (AD) methods emerged as an appropriate solution to detect these anomalous failure states but do not distinguish failures from benign deviations.
We introduce \textbf{FIDeL} (Failure Identification in Demonstration Learning), a policy-independent failure detection module. Leveraging recent AD methods, FIDeL builds a compact representation of nominal demonstrations and aligns incoming observations via optimal transport matching to produce anomaly scores and heatmaps. Spatio-temporal thresholds are derived with an extension of conformal prediction, and a Vision–Language Model (VLM) performs semantic filtering to discriminate benign anomalies from genuine failures. We also introduce \textbf{BotFails}, a multimodal dataset of real-world tasks for failure detection in robotics. FIDeL consistently outperforms state-of-the-art baselines, yielding +5.30\% AUROC in anomaly detection and +17.38\% failure-detection accuracy on BotFails compared to existing methods. Videos of FIDeL can be found on our website : \\\textcolor{darkorange}{\url{https://cea-list.github.io/FIDeL/}}

\end{abstract}

%% file: sec/1_intro.tex
\section{INTRODUCTION}

Recent progresses in imitation learning (IL)~\cite{pizeropointcinq, diffusionpolicy, zhao2023learningfinegrainedbimanualmanipulation} have the potential to significantly enhance the flexibility and adaptability of robotic systems. Unfortunately, current learned policies cannot be deployed in real world environments because their behavior is not defined when they are in a situation that is not part of the training set. Such events are unavoidable in factories or human-centered environments, arising from defective objects, operator mistakes, or environmental shifts, and they cannot be exhaustively anticipated in training data.

A prevalent strategy is to assume that the training dataset is sufficiently diverse to encompass all possible scenarios. Implicitly, this amounts to presuming that no situation is off distribution. While this assumption simplifies the problem, it is unrealistic in practice, many rare events will never appear in training data. The second approach is to design models that can quantify the uncertainty of their decision, for instance with Bayesian models~\cite{Gal2016Dropout, Kendall2017WhatUncertainties}.
However, uncertainty quantification remains an open challenge in IL and RL~\cite{lockwood2022reviewuncertaintydeepreinforcement,Zhu2019UQRL}, and is not integrated into the highest-performing learning algorithms~\cite{pizeropointcinq, diffusionpolicy, zhao2023learningfinegrainedbimanualmanipulation}. A more practical direction is to rely on an independent monitoring module that evaluates execution at runtime~\cite{roth2022,fan2018videoanomalydetectionlocalization,hafezSafeLLMControlledRobots2025,agia2024STAC,xu2025FailDetect}.

In this paper, we build on recent work~\cite{roth2022, fan2018videoanomalydetectionlocalization, hafezSafeLLMControlledRobots2025, agia2024STAC, xu2025FailDetect} and propose an independent failure monitor, designed to detect situations in which a policy should be interrupted. We frame the problem in terms of Anomaly Detection (AD). Anomalies correspond to deviations from the expected distribution, which may—but do not always—indicate failure states. Our approach leverages the One-Class (OC) paradigm, a well-established method in computer vision~\cite{Cui_2023, abdalla2024videoanomalydetection10, wu2024deeplearningvideoanomaly, chalapathy2019deeplearninganomalydetection, tax1999support}. In OC learning, a model is trained solely on data from a “normal” class, and any significant deviation from this baseline is flagged as anomalous. %Although OC learning can be leveraged to supervise any robot task execution (learnt policy or not) provided a database of nominal execution samples, 
This formulation is particularly well-suited to IL, where the space of possible deviations is unbounded, yet expert demonstrations naturally provide a reliable definition of normality. Thus, we treat expert demonstrations as our baseline and regard any deviation as an anomaly—yielding an implicit and comprehensive definition of undesirable behavior.

%There is however a large domain gap between state-of-the-art Vision AD (VAD) and robotics. Indeed, the nature of the visual scenes differs substantially beetween existing VAD and robotics. VAD research typically focuses on specific applications--crowd scenes~\cite{luo2017revisit, abnormal2013lu, mahadevan2010anomaly}, industrial assembly lines~\cite{bergmann2021mvtec, bergmann2022beyond, zou2022spotthedifferenceselfsupervisedpretraininganomaly}, while robotics encompasses a wide range of tasks, resulting in a much broader spectrum of potential anomalies. 

That said, the notion of anomaly encompasses a broader range of situations than failure. Not all anomalies correspond to failure-related events: some may be entirely benign—for instance, a fly landing on a table or an object slightly shifting in the background—and should not trigger unnecessary or costly interventions. This mismatch reveals a key limitation of existing AD methods: while they can reliably highlight deviations from nominal behavior, they lack the ability to discriminate between harmless anomalies and those that truly compromise task execution. In robotics, this distinction is essential. What is required is a mechanism that can semantically interpret anomalies and explicitly determine whether they represent benign variations or genuine failures.

To bridge the gap between traditional Vision Anomaly Detection and the unique demands of robotic applications, we introduce \textbf{FIDeL} (\textbf{F}ailure \textbf{I}dentification in \textbf{De}monstration \textbf{L}earning), a failure detection module designed to complement IL policies in robotics. \textbf{FIDeL} performs post-hoc \textbf{anomaly detection} by comparing new observations with stored representations of normal demonstrations using Optimal Transport (OT). To decide whether or not an AD score indicates an anomaly, we leverage a \textbf{Conformal Prediction} (CP)–based thresholding~\cite{lei2017distributionfreepredictiveinferenceregression, diquigiovanni2021importancebandfinitesampleexact} mechanism, which accounts for both temporal variations in anomaly scores and spatial variations across visual patches. When the anomaly score exceeds the CP threshold, \textbf{FIDeL} triggers a \textbf{semantic filter} based on a vision-language model (VLM), which determines whether the detected anomaly is failure-related or benign. This three-stage process allows for efficient failure detection while reducing false positives, thus preserving productivity.

%\quentinrmk{Ajouter que notre modèle n'est pas forcément pensé pour être le seul mécanisme de sécurité. L'utilisation d'un VLM implique une certaine instabilité et imprévisibilité qui demande des mécanismes conjoints. Notamment des modèles de détection supervisés qui sont entraînés pour détecter des situations à risque spécifiques, des modèles de sécurité classiques... ctrl barrier function etc... L'utilité de notre algo c'est sa capacité à détecter des situations unsafe \textbf{incongrues}, qui n'ont pas été prévues}
%%%%

%\quentinrmk{Il faudrait aussi ajouter ça mais je ne vois pas encore bien comment l'incorporer : concerns about anomaly detection being reactive.While our method detects anomalies only once they begin to manifest, this still allows for timely and practical intervention. In real-world scenarios, early interruption after onset can prevent harm — e.g., stopping a robot about to solder when a hand enters its workspace, or halting a spill to avoid waste or short circuits. Crucially, anomalies are detected on input observation, before the next IL policy action is executed, allowing the system to interrupt behavior and prevent further unsafe outcomes. Thus, detection at onset is often sufficient to avoid harm. Moreover, our system is compatible with upstream monitors and can serve as a last line of defense.}\fabricermk{c'est le bon discours, je pense qu'il faut un paragraphe dans l'intro, près de là où on mentionne le terme post-hoc, ou on dit où temporellement dans le processus notre module s'intègre}

Our contributions are summarized as follows:  
\begin{itemize}  
    \item Following the One-Class paradigm, we design a \textbf{novel Vision Anomaly Detection, representation-based algorithm} tailored for autonomous robotics.  
    \item We extend Conformal Prediction to handle both \textbf{temporal and spatial variations}, with memory-based temporal alignment that ensures robustness to variable execution speeds.  
    \item We propose a \textbf{semantic filtering mechanism} using a VLM and anomaly heatmaps, which discriminates between benign anomalies and task-critical failures.  
    \item We enhance interpretability through \textbf{localized heatmaps and VLM-based explanations}, providing human-readable justifications for failure detection.  
    \item We present \textbf{BotFails} a \textbf{dedicated training and evaluation dataset} collected on LeRobot~\cite{cadene2024lerobot} for benchmarking anomaly and failure detection in robotics, addressing the scarcity of failure-oriented datasets.  
    \item We demonstrate significant performance gains over baselines in both anomaly detection ($+5.30\%$ AUROC) and failure detection tasks ($+17,38\%$ accuracy).  
\end{itemize}

%We also provide a supplementary video showcasing qualitative results, highlighting how our method localizes anomalies, applies semantic filtering, and triggers failure detection in real robotic scenarios.  

%% file: sec/2_related_work.tex
\section{RELATED WORK}

\subsection{Vision Anomaly Detection}

Vision AD is a well-established problem in computer vision, particularly within the unsupervised learning paradigm. Existing methods can be broadly categorized into four families~\cite{Cui_2023}:
\textbf{(1) Augmentation-based methods} add artificial anomalies to normal samples and train a classifier to recognize them~\cite{li2021cutpaste, zavrtanik2021draem, schluter2022}. However, we found no straightforward way to adapt these approaches to robotics, primarily due to the challenge of generating realistic anomalies in robot and object motions. 
%In particular, ensuring that the synthesized anomalies are both non-trivial to detect and meaningfully out-of-distribution relative to expert demonstrations presents a significant difficulty.
\textbf{(2) Reconstruction-based methods} learn to reconstruct normal inputs, using the reconstruction error as an anomaly score. Common architectures include Autoencoders (AE), Variational Autoencoders (VAE) or Generative Adversarial Networks (GANs)\cite{venkataramanan2020VAEandGAN, liu2021AEandGAN, Shi_2021AE}, and Student-Teacher frameworks\cite{wang2021studentteacherfeaturepyramidmatching, yamada2022studentteacherdiscriminativenetworks, rudolph2022asymmetricstudentteachernetworksindustrial}. However, such models often generalize well enough to reconstruct anomalous inputs, thus failing to highlight anomalous regions effectively~\cite{10208652, bouman2025autoencodersanomalydetectionunreliable}.
\textbf{(3) Normalizing Flows} (NF)~\cite{rezende2016NF} based methods~\cite{gudovskiy2021NF, rudolph2021NF, yu2021fastflow} learn an invertible transformation from the data distribution to a well-defined prior, typically a Gaussian. Anomalies are detected when the likelihood of a test sample under the learned distribution deviates significantly from the normal class. Despite their theoretical appeal, NF exhibit “likelihood paradoxes,” often assigning higher likelihood to anomalous images than normal ones~\cite{kirichenko2020flowood}.
\textbf{(4) Representation-based methods} \cite{defard2020padim, roth2022, Wang2021, zheng2022} use pretrained, frozen networks to extract semantic embeddings and compute anomaly scores based on their distance to normal class features. Early methods like PaDiM\cite{defard2020padim} model per-patch Gaussian distributions, while PatchCore~\cite{roth2022} retrieves outliers from a memory bank of normal features. These approaches are particularly appealing for robotics as they are sample-efficient and interpretable, providing localized heatmaps that highlight anomalous regions~\cite{defard2020padim, roth2022}. Their efficiency and explainability make them especially suited for failure detection. We evaluate three AD families (see Section~\ref{sec:evaluation}).

\subsection{Failure detection in autonomous robotics}

Ensuring that autonomous robotic agents can detect and respond to abnormal behaviors or failure during deployment is critical for both performance and human trust.

Traditional approaches to failure detection have focused on detection through supervised learning frameworks that rely on labeled examples of failures~\cite{liu2023modelbasedruntimemonitoringinteractive, gokmen2023askinghelpfailureprediction, liuMultiTaskInteractiveRobot2024, guptaDetectingMitigatingSystemLevel2024}. For instance, Liu et al.~\cite{liu2023modelbasedruntimemonitoringinteractive} trained an LSTM-based classifier using latent embeddings from a conditional VAE for RNN-based BC policies, while Gokmen et al.~\cite{gokmen2023askinghelpfailureprediction} employed a jointly trained state-value function to anticipate failure states in behavior cloning setups. However, these methods require failure data trajectories for training, limiting their applicability in novel or unstructured environments.

Several alternative approaches have been explored. Wang et al.~\cite{wang2024groundinglanguageplansdemonstrations} collect failure data via self-reset to train a classifier, requiring around 2,000 trajectories (2 hours of data), which limits scalability. He et al.~\cite{he2024rediffuser} use random network distillation to filter out out-of-distribution trajectories, while Sun et al.~\cite{sun2023conformal} reduce uncertainty by predicting reward intervals. However, these methods do not address failure detection during execution. In contrast, other approaches such as Hafez et al.~\cite{hafezSafeLLMControlledRobots2025} verify LLM-generated trajectories using reachability analysis, and Agia et al.\cite{agia2024STAC} introduce a statistical temporal action consistency (STAC) metric with VLMs to detect execution-time anomalies. Still, both focus solely on trajectory-level data.

The closest work to ours, FAIL-Detect~\cite{xu2025FailDetect}, performs runtime failure detection in imitation learning with Continuous Normalizing Flows (CNF)~\cite{chen2019neuralordinarydifferentialequations}, but does not distinguish benign anomalies from task-critical failures and assumes that the task is performed with the same temporal consistency during inference as in demonstrations. Instead, our method relies on a representation based formulation that captures both temporal progression and spatial structure, enabling alignment across variable execution speeds, localized heatmaps, and more interpretable scores. Combined with a semantic filtering stage using a VLM, this design narrows detection to genuine failures and yields significant gains, achieving +17.38\% accuracy on BotFails.

%% file: sec/3_method.tex
\begin{figure*}[t]
    \centering
    \includegraphics[width=\textwidth]{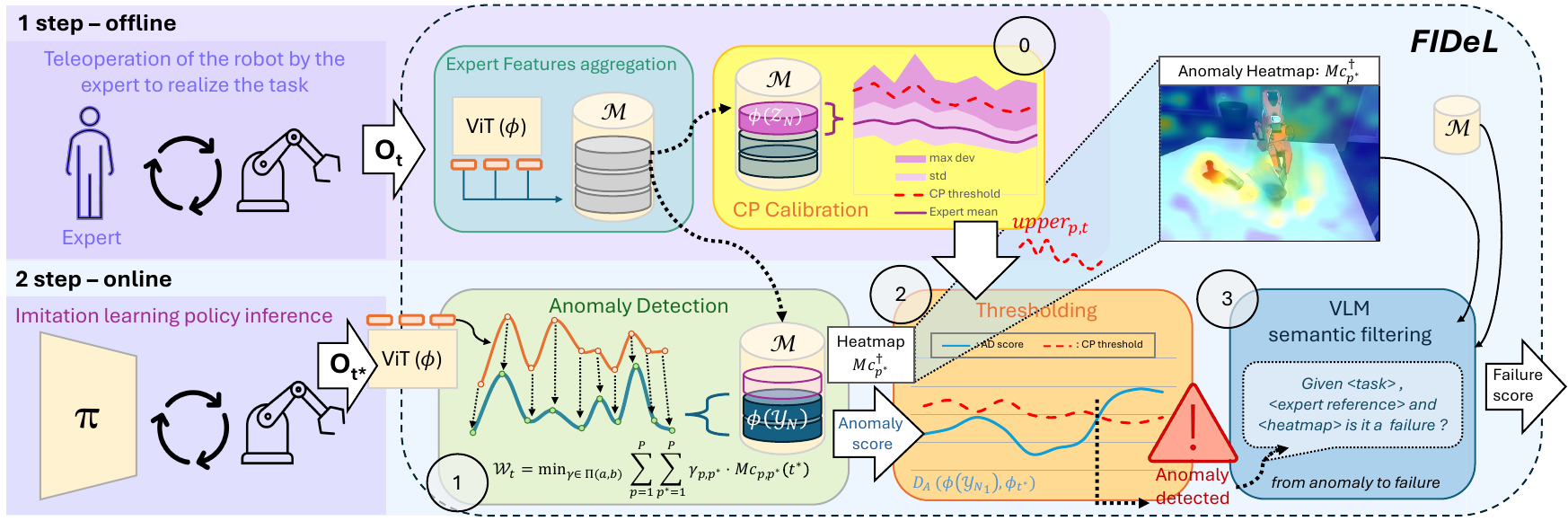}
    \caption{
    We introduce \textbf{FIDeL}, a framework for detecting failures in imitation learning (IL) policies.  
    \textbf{- Offline -}  
    Expert demonstrations are first encoded and stored in a memory buffer $\mathcal{M}$.  
    \textcolor{darkyellow}{0. Conformal Prediction Calibration} — A decision threshold is computed from $\mathcal{M}$ using Conformal Prediction to determine when a score should be considered anomalous.  
    \textbf{- Online -}  
    \textcolor{darkgreen}{1. Anomaly Detection} — During policy execution, anomaly scores and heatmaps are computed from incoming observations.  
    \textcolor{darkorange}{2. Thresholding} — If the score exceeds the calibrated threshold, an anomaly is flagged.  
    \textcolor{darkblue}{3. Semantic Filtering} — Since not all anomalies are failure-related, a VLM filter discards benign deviations and flags only failures that threaten task success.  
    }
    \label{fig:global}
\end{figure*}

\section{Method}  
\label{sec:method}  

\subsection{Problem formulation}  
\label{subsec:problem}  

We consider a generative robotic policy $f_\theta$ trained using Imitation Learning. At inference, the policy maps an observation $O_t$ to an action $A_t = f_\theta(O_t)$. Our goal is to design a monitoring module capable of identifying execution anomalies and, among them, failures that compromise task completion.  
We decompose the problem into two levels: \textbf{Anomaly detection}: an anomaly detector $D_A$ assigns a score to each observation $O_t$, reflecting the likelihood of deviation from nominal behavior. \textbf{Failure detection}: a failure detector $D_F$ refines this decision by filtering out benign anomalies (e.g., harmless scene changes) while flagging failure-related anomalies.  
We adopt a one-class formulation, requiring only nominal demonstrations. Let $\mathcal{X}_N$ denote the dataset of expert trajectories, consisting of at most a few dozen episodes. It is randomly partitioned into two subsets of equal sizes: $\mathcal{Y}_N$ for constructing the anomaly detector $D_A$ and $\mathcal{Z}_N$ for calibrating thresholds. 

\subsection{Method overview}  
\label{subsec:overview}  

Our framework comprises two stages: an \textbf{Offline phase} during which observations from $\mathcal{X}_N$ are encoded using a vision encoder $\phi$, yielding patch-level features $\phi_t = \phi(O_t)$. These features are aggregated into a statistical memory $\mathcal{M}$ (Sec.~\ref{subsec:memory}), which models the nominal distribution. The calibration subset $\mathcal{Z}_N$ is then used to derive adaptive thresholds via conformal prediction, producing time and patch dependent bounds. During the \textbf{online phase.}, at each timestep $t^*$, the policy outputs an action $A_{t^*} = f_\theta(O_{t^*})$. In parallel, as illustrated in Fig.~\ref{fig:global}, the anomaly detector $D_A$ computes a score $D_A(\phi(\mathcal{Y}_N), \phi_{t^*})$ by aligning query features $\phi_{t^*}$ with keys from $\mathcal{Y}_N$ using optimal transport. This score is compared against conformal thresholds to decide if $O_{t^*}$ is anomalous. If flagged, the failure detector $D_F$ invokes a VLM to distinguish harmless anomalies from genuine failures.  

\subsection{Memory representation}  
\label{subsec:memory}  

The memory module $\mathcal{M}$ is built offline from nominal demonstrations. After encoding trajectories using a vision encoder (resnet18~\cite{he2015deepresiduallearningimage} or dinoV2~\cite{oquab2024dinov2learningrobustvisual}), we obtain a feature tensor $\phi(\mathcal{Y}_N) \in \mathbb{R}^{N \times T \times P \times F}$, where $N$ is the number of episodes, $T$ the horizon length, $P$ the number of spatial patches per frame, and $F$ the feature dimension. To reduce memory cost and ensure efficient inference, features are aggregated across episodes by computing Gaussian statistics. The resulting memory encodes the expected distribution of nominal features:  
\[
\mathcal{M} = \left\{ \mu_{t,p,f}, \, \sigma_{t,p,f} \;\middle|\; t \in \llbracket 1, T \rrbracket, \, p \in \llbracket 1, P \rrbracket, \, f \in \llbracket 1, F \rrbracket \right\},
\]  
where $\mu_{t,p,f}$ and $\sigma_{t,p,f}$ are the empirical mean and standard deviation of feature dimension $f$, estimated across episodes for patch $p$ at timestep $t$.

\begin{figure*}[!t]
    \centering
    \includegraphics[width=\textwidth]{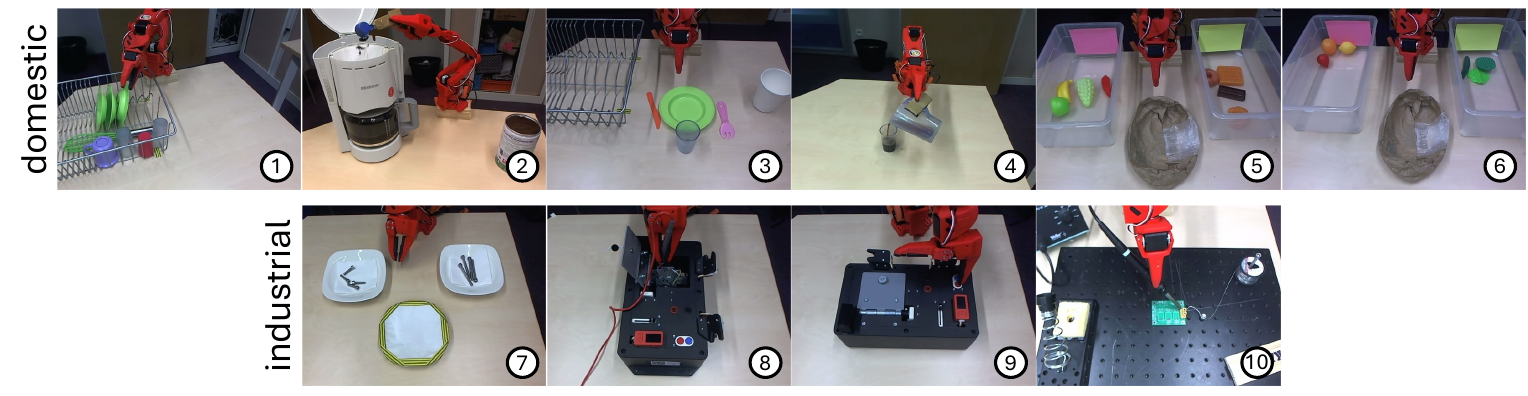}
    \caption{\textbf{BotFails dataset illustration} - all images are illustrations of BotFails tasks executed by the expert - \textbf{Domestic} - 1. clear away the dishes, 2. make coffee, 3. set the table, 4. pour coffee, 5. sort groceries, 6. sort fruits and vegetables - \textbf{Industrial} - 7. sort screws, 8. measure voltage, 9. press buttons, 10. solder.
    }
    \label{fig:botfails}
\end{figure*}

\subsection{Anomaly score computation}

At runtime, in parallel with the operation of the policy \( f_{\theta} \), we retrieve the observation \( O_{t^*} \). Features are then extracted from \( O_{t^*} \) using the vision encoder, resulting in \( \phi_{t^*}^{query} = \phi(O_{t^*}) \in \mathbb{R}^{P \times F} \). Subsequently, we compare this query data--whose indices hold a "$*$"-- with the memory, i.e., the stored data, which we refer to as the keys. 
Anomaly scoring relies on an optimal transport (OT) alignment between the query and memorized features~\cite{papagiannis2022imitationlearningsinkhorndistances, dadashi2021primalwassersteinimitationlearning}.
Rather than comparing patches one by one, OT seeks the best global alignment between the two sets of patches: how much "feature mass" from query patches needs to be moved to match the key (memory) patches, and at what cost. Formally, we look for a transport plan $\gamma \in \mathbb{R}_+^{P*P}$ that redistributes the query mass into the key mass with minimum total cost.
We define a ground cost function 
$c: \mathbb{R}^F \times \mathbb{R}^F \to \mathbb{R}^+$ 
between two feature vectors, implemented here as normalized Euclidean distance (i.e., a diagonal Mahalanobis distance), where each feature dimension is scaled by the corresponding standard deviation stored in $\mathcal{M}$. This gives rise to a cost matrix $M_c(t) \in \mathbb{R}^{P \times P}$ for each memory timestep $t$:  
\begin{equation}
M_{c}(t)[p,p^*] = c\big(\phi^{query}_{t^*,p^*}, \, (\mu_{t,p}, \sigma_{t,p}) \big). 
\label{eq:Mc}
\end{equation}
We then seek a coupling matrix $\gamma \in \Pi(a,b)$, where 
\[
\Pi(a,b) = \big\{ \gamma \in \mathbb{R}_+^{P \times P} \;\big|\; \gamma \mathbf{1} = a, \, \gamma^\top \mathbf{1} = b \big\}
\]
is the set of admissible transport plans between two uniform distributions 
$a, b \in \Delta^P$.  
The OT cost between query and memory features at time $t$ is thus:  
\begin{equation}
\mathcal{W}_t = \min_{\gamma \in \Pi(a,b)} \langle M_c(t), \gamma \rangle
\end{equation}
where $\langle \cdot,\cdot\rangle$ denotes the Frobenius inner product.  

The final anomaly score is obtained as the minimum transport cost across nominal timesteps:  
\begin{equation}
D_A(\phi(\mathcal{Y}_N), \phi_{t^*}) = \min_{t \in \llbracket 1,T \rrbracket} \mathcal{W}_t ,\quad
t_{\min} = \arg\min_{t \in \llbracket 1,T \rrbracket} \mathcal{W}_t
\label{eq:anomaly_score}
\end{equation}

In practice, solving the OT problem exactly is costly; we rely on entropic regularization (Sinkhorn iterations) to compute an efficient approximation~\cite{papagiannis2022imitationlearningsinkhorndistances}.

\subsection{Thresholding via Spatially-Aware Conformal Prediction}
\label{sec:threshold}  

% =======================================================

We extend the Conformal Prediction (CP) framework~\cite{lei2017distributionfreepredictiveinferenceregression, diquigiovanni2021importancebandfinitesampleexact} to define a dynamic threshold for our anomaly detector $D_A$, accounting for both temporal structure and patch-level variations.

\begin{figure*}[!h]
    \centering
    \includegraphics[width=\textwidth]{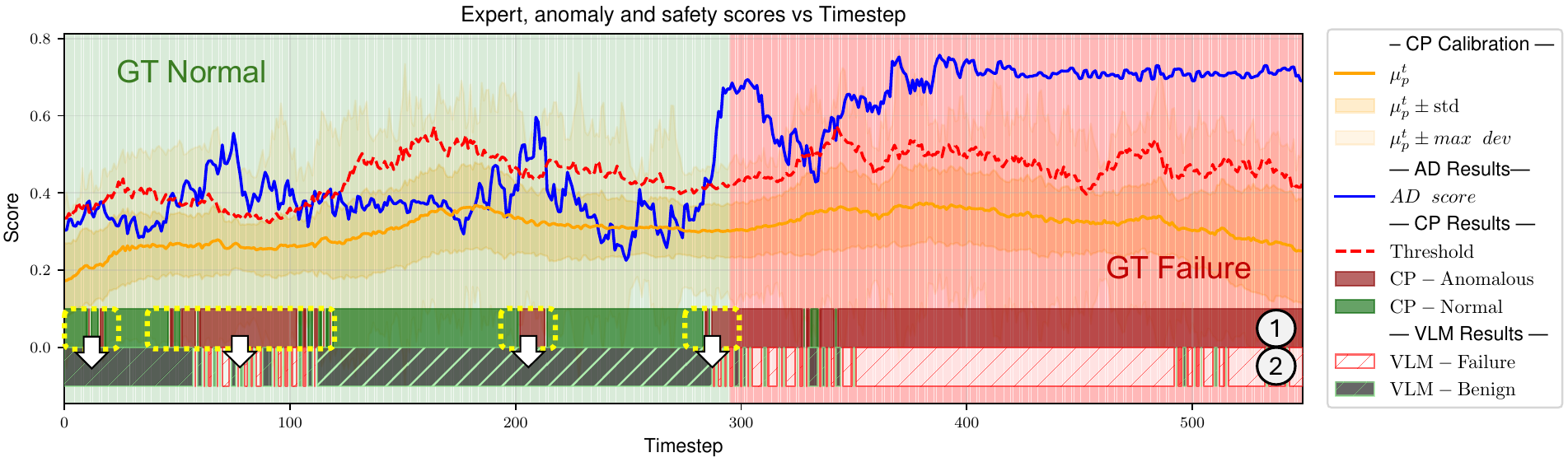}
    \caption{\textbf{Score illustration} - Real-$\pi$ soldering - episode 15 over 20 episodes of the evaluation set, score obtained with Representation and CP time. The \textcolor{darkblue}{Anomaly Detection score} corresponds to the output of the AD module $D_A(\phi(\mathcal{Y}_N), \phi_{t^*})$. \textcolor{darkyellow}{$\mu^t_p$} is obtained when computing $D_A(\phi(\mathcal{Y}_N), \phi(\mathcal{E}_A))$ and allows to compute the \textcolor{red}{Threshold values}, see section \ref{sec:threshold}. When the \textcolor{darkblue}{Anomaly Detection score} is below the \textcolor{red}{Threshold}, no anomaly is detected (\textcolor{darkgreen}{CP-Normal}) and when it is above, an anomaly is flagged (\textcolor{darkred}{CP-Anomalous}). 
    We observe that transitioning from anomaly detection \textcircled{\tiny{1}} to failure detection \textcircled{\tiny{2}} through the use of the VLM reduces the number of false positives (highlighted in the \textcolor{yellow}{yellow} boxes). However, this additional filtering step also introduces a small number of false negatives.
    }
    \label{fig:score_illustration}
\end{figure*}

\textbf{- Offline calibration -} \\
We divide the expert calibration set $\mathcal{Z}_N$ in two subsets $\mathcal{E}_A$ and $\mathcal{E}_B$. Then, we compute the anomaly calibration score $D_A(\phi(\mathcal{Y}_N), \phi(\mathcal{E}_A))$ which is used to compute the mean anomaly score $\mu^t_p$ over nominal examples, while $D_A(\phi(\mathcal{Y}_N), \phi(\mathcal{E}_B))$ is used to compute deviations. The split aims to maintain the exchangeability assumption~\cite{lei2017distributionfreepredictiveinferenceregression}.
A threshold bandwidth $h$ is then set as the $(1 - \alpha)$-quantile of the maximum deviations observed on $\mathcal{E}_B$. This bandwidth yields a spatio-temporal upper bound $\mathrm{upper}_{p,t} = \mu^t_p + \rho^t_p . h$, which controls the false positive rate at level $\alpha$ under exchangeability. With $\rho^t_p$, a patch-wise modulation factor, capturing variability across time and space~\cite{lei2017distributionfreepredictiveinferenceregression}. Figure~\ref{fig:score_illustration} depicts the upper bound (Threshold) as a dotted red line.

\textbf{- Online thresholding -} \\
At runtime, the current observation is aligned to its closest nominal timestep $t_\text{min}$ using OT (Eq.~\eqref{eq:anomaly_score}), and patch-level scores are compared to their corresponding bounds $\mathrm{upper}_{p^*,t_\text{min}}$. An input is flagged as anomalous when the fraction of patches exceeding their respective bounds surpasses a user-defined percentage. This parameter allows tuning the detector’s sensitivity to the task context: stricter values improve responsiveness in fine-grained or sensitive scenarios, while more permissive ones reduce false positives in less sensitive settings. Additionally, non-critical spatial regions can be masked, further focusing detection on areas of interest.

% =========================

\subsection{Semantic filtering}

We leverage Qwen 2.5-7b~\cite{qwen2025qwen25technicalreport} as a Vision-Language Model (VLM) to distinguish between benign anomalies and task-relevant failures. Formally, we define the failure detector $D_F$ as a binary classifier operating on both contextual task information and visual evidence:  
\[
D_F : \big( d,\, O(\mathcal{X}_{N,t_{\min}}),\, O_{t^*},\, Mc^{\dagger}_{p^*} \big) \;\mapsto\; \{0,1\},
\]
where $d$ is a task-specific textual description provided by the user, $O(\mathcal{X}_{N,t_{\min}}) \in O(\mathcal{X}_N)$ is a reference (expert) demonstration frame at timestep $t_{\min}$ (see Eq.~\eqref{eq:anomaly_score}), $O_{t^*}$ is the current observation, and $Mc^{\dagger}_{p^*}$ is the anomaly heatmap highlighting the most suspicious region(s) in the image.  
$Mc^{\dagger}_{p^*}$ $\in \mathbb{R}^{P}$ is derived from the cost matrix (eq.(\ref{eq:Mc})) at the most similar time step \( t_{\text{min}} \) (eq.(\ref{eq:anomaly_score})) by extracting the minimum transport cost per query patch:
\begin{equation}
Mc^{\dagger}_{p^*} = \min_{p} \left( Mc_{p, p^*}(t_{\text{min}}) \right)
\label{eq:heatmap}
\end{equation}
The VLM is prompted with $(d, O(\mathcal{X}_{N,t_{\min}}), O_{t^*}, Mc^{\dagger}_{p^*})$ and asked to:  
\begin{enumerate}
    \item Identify the semantic nature of the anomaly by comparing the two images and interpreting the heatmap.  
    \item Classify the anomaly as either:  
\textbf{False positive} ($D_F=0$): the anomaly is benign, the robot may continue.  
\textbf{Failure} ($D_F=1$): the anomaly is a failure.  
\end{enumerate}

The combination of nominal reference ($O(\mathcal{X}_{N,t_{\min}})$) with structured context ($d$) and localized anomaly cues ($Mc^{\dagger}_{p^*}$), allows to constrain the reasoning space of the VLM. 
%We include the VLM prompt and answers in the supplementary video.

% ==================================

\begin{figure*}[!t]
    \centering
    \includegraphics[width=\textwidth]{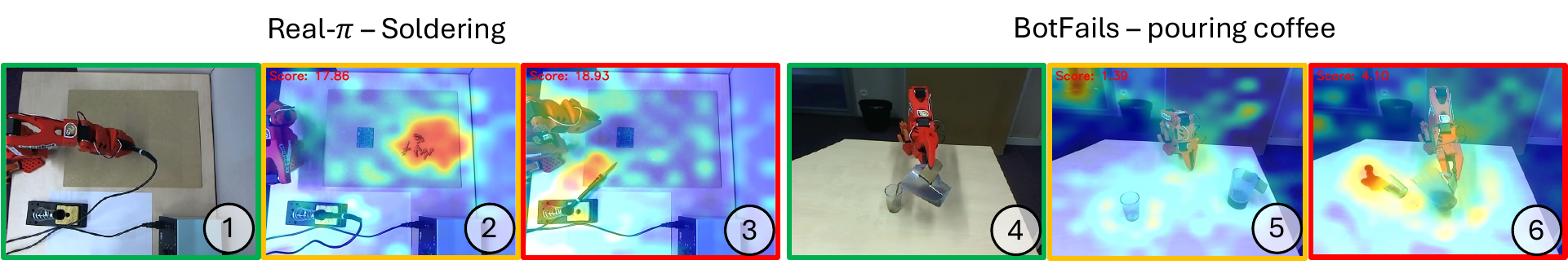}
    \caption{\textbf{Heatmaps illustration} - obtained using Representation AD and temporal/spatial CP - \textbf{Real-$\pi$}: \textcircled{\tiny{1}} expert, \textcircled{\tiny{2}} benign anomaly: screws present in the work plan, \textcircled{\tiny{3}} failure: dropping the iron - \textbf{BotFails}: \textcircled{\tiny{4}} expert, \textcircled{\tiny{5}} benign anomaly: someone walking in the background (top left corner), \textcircled{\tiny{6}} failure: spilling the cup.
    }
    \label{fig:heatmaps}
\end{figure*}

%% file: sec/4_evaluation.tex
% ===============================================================================================
% table for AD evaluation

\begin{table*}[h]
    \centering
    \resizebox{\textwidth}{!}{%
    \begin{tabular}{l c c c c c c c c c c}
        \toprule
        & \multicolumn{2}{c}{\textbf{Representation (ours)}} 
        & \multicolumn{2}{c}{\textit{logpZ0}~\cite{xu2025FailDetect}} 
        & \multicolumn{2}{c}{\textit{lopO}}
        & \multicolumn{2}{c}{\textit{AE}} 
        & \multicolumn{2}{c}{\textit{STAC}~\cite{agia2024STAC}} \\
        
        & AUROC $\uparrow$ & F1@Opt $\uparrow$
        & AUROC $\uparrow$ & F1@Opt $\uparrow$
        & AUROC $\uparrow$ & F1@Opt $\uparrow$
        & AUROC $\uparrow$ & F1@Opt $\uparrow$
        & AUROC $\uparrow$ & F1@Opt $\uparrow$ \\
        
        \midrule
        
        \multirow{2}{*}{\textbf{BotFails}}
        
        & \cellcolor{green!20} \boldmath{$70.50$} & \cellcolor{green!20} \boldmath{$62.90$}
        & $65.20$ & $59.87$
        & $58.30$ & $53.48$
        & $49.90$ & $52.12$ 
        & N/A & N/A \\
        
        & \cellcolor{green!20} \boldmath{$\pm 0$} & \cellcolor{green!20} \boldmath{$\pm 0$} & $\pm 0.15$ & $\pm 0.06$ & $\pm 0.70$ & $\pm 0.02$ & $\pm 1.08$ & $\pm 0.08$ & N/A & N/A \\
        
       \cmidrule(lr){1-1} \cmidrule(lr){2-3} \cmidrule(lr){4-5} \cmidrule(lr){6-7} \cmidrule(lr){8-9} \cmidrule(lr){10-11}

        \multirow{2}{*}{\textbf{Real-$\pi$ soldering}} 
        
        & \cellcolor{green!20} \boldmath{$86.72$} & \cellcolor{green!20}\boldmath{$78.29$}
        & $82.62 $ & $78.03 $
        & $76.16 $ & $70.99 $ 
        & $81.07 $ & $76.33 $
        & $53.55 $ & $63.93 $ \\
        
        & \cellcolor{green!20}\boldmath{$\pm 0$} & \cellcolor{green!20}\boldmath{$\pm 0$} & $\pm 0.08$ & $\pm 0.02$ & $\pm 0.42$ & $\pm 0.03$ & $\pm 0.88$ & $\pm 0.06$ & $\pm 0$ & $\pm 0$ \\
        
        \bottomrule
    \end{tabular}
    } % end resizebox
    \caption{\textbf{Anomaly Detection evaluation before thresholding} – mean AUROC and mean F1-score at optimal threshold (\textbf{in \%}) for anomaly detection only evaluated on the \textbf{BotFails} dataset and \textbf{Real-$\pi$ soldering task}, across various AD methods. Best results are highlighted in green and bold. Each task score reflects performance over multiple rollouts (several dozens per task).}
    \label{tab:AD_results}
\end{table*}

\section{EVALUATION}
\label{sec:evaluation}

To rigorously evaluate the effectiveness of our failure detection approach, we introduce a dedicated dataset specifically designed for robotic failure monitoring. Our method is benchmarked on this dataset alongside several state-of-the-art baseline methods adapted to our experimental setting.

\subsection{Dataset}
\subsubsection{We introduce \textbf{BotFails}} given the scarcity of available datasets in the robotics failure detection field, we created a new dataset specifically designed for general failure situations in robotics, incorporating vision, proprioception, and natural language instructions. Data collection was performed using master arm teleoperation with LeRobot~\cite{cadene2024lerobot}, a real-world robotic platform. BotFails has been curated to maximize semantic diversity across task types. The dataset covers 10 distinct tasks, with 6 tasks reflecting domestic environments, and 4 tasks representative of industrial contexts. The tasks are illustrated in Figure \ref{fig:botfails}. Each task features several anomaly types. Some are not failure-related (e.g. an unknown object is present in the scene) and some are real failures (e.g. manipulation mistake, semantic mistake...).

\subsubsection{\textbf{Real-\boldmath{$\pi$}} dataset}
We additionally evaluated FIDeL on more realistic data by performing inference on trajectories generated with ACT~\cite{zhao2023learningfinegrainedbimanualmanipulation}, an IL policy. While these data contain fewer types of anomalies/failures—due to the lack of control over the policy’s behavior—they better reflect real-world execution scenarios. We conducted experiments on a soldering task. Two types of labels are available with these datasets: anomaly detection labels and failure labels which only include anomalies related to failure scenarios.

\subsection{Baselines}

We evaluate FIDeL against three families of baselines: (i) AD methods relying exclusively on nominal data, (ii) thresholding strategies for anomaly scoring, and (iii) end-to-end failure detection modules that combine detection with additional filtering or semantic reasoning. 

\subsubsection{Anomaly Detection (AD) baselines}

We first compare against representative approaches from recent advances in robotic AD and Vision Anomaly Detection (VAD). Like FIDeL, these methods are trained exclusively on nominal demonstrations, without requiring annotated anomalies.  

\textbf{FAIL-Detect}~\cite{xu2025FailDetect} introduces two anomaly scoring mechanisms based on Conditional Normalizing Flows (CNFs): \textbf{lopO} and \textbf{logpZ0}~\cite{xu2024normalizingflowneuralnetworks}.  
- \textbf{lopO} estimates the likelihood of an observation $O_t$ under a CNF fitted on expert demonstrations; anomalies correspond to low-likelihood samples.  
- \textbf{logpZ0}, instead, integrates the CNF backward from $O_t$ to compute its corresponding latent variable $Z_{O_t}$. Under nominal conditions, $Z_{O_t}$ is approximately Gaussian, so large values of $\|Z_{O_t}\|^2$ are indicative of anomalies. Compared to lopO, this method avoids explicitly computing the flow’s divergence, improving stability in high-dimensional settings.  

%Since no official implementation was released, we reimplemented both variants following the hyperparameters and training protocol of the original paper. As in their setup, we use a ResNet-18~\cite{he2015deepresiduallearningimage} encoder, and define $O_t$ as the concatenation of the two most recent RGB frames and the associated robot joint states.  

In addition, we include a \textbf{reconstruction-based AD baseline}, inspired by SOTA methods for video anomaly detection~\cite{Shi_2021AE, hasan2016learningtemporalregularityvideo, Wang2018}. We train an AutoEncoder (AE) on expert data to learn a compact latent representation. At inference, high reconstruction error signals a deviation from expert-like behavior and is used as an anomaly score.  

Finally, we consider STAC (Statistical Temporal Action Consistency)~\cite{agia2024STAC} which monitors the temporal consistency of a generative policy. At each step, the policy predicts a sequence of future actions; overlapping parts of successive predictions are compared using statistical distances (e.g., MMD, KL). When the policy is in-distribution, these overlapping action distributions remain consistent, yielding low distances. Conversely, large divergences signal out-of-distribution states and potential failure. We can only evaluate STAC on the Real-$\pi$ dataset given that it requires the sequence of actions predicted by the policy.

\subsubsection{Thresholding baselines}

Raw anomaly scores need to be mapped to binary anomaly decisions. To assess the contribution of our conformal thresholding mechanism, we compare against several thresholding strategies:  
- \textbf{CP-time}, a standard conformal prediction approach using only temporal deviations from reference demonstrations.  
- \textbf{CP-time+space}, our spatial extension that accounts for both temporal and spatial (patch-level) variations.  
- \textbf{Gaussian assumption}, a simpler baseline with thresholds derived from a Gaussian fit of calibration scores (mean and variance) at each timestep.  

\subsubsection{End-to-end failure detection baselines}

Beyond isolated AD modules, we also benchmark FIDeL against SOTA full failure detection pipelines:  
- \textbf{FAIL-Detect}~\cite{xu2025FailDetect}, in its end-to-end configuration.  
- \textbf{Sentinel}~\cite{agia2024STAC}, which combines STAC and VLM monitoring in parallel.  
- \textbf{VLM only}: we use Qwen 2.5~\cite{qwen2025qwen25technicalreport}, as a failure classifier. It receives the 5 last images and the user prompt and decides whether a failure occurred or not.

\subsection{Evaluation Protocol}

\textbf{Anomaly Detection (AD) module.}
We first evaluate anomaly scoring independently of thresholds by extracting raw anomaly scores. Labels are defined at the frame level (0 = nominal, 1 = anomaly). Metrics: AUROC (balanced evaluation of TPR/FPR across thresholds) and F1 at the optimal threshold (best precision–recall balance).

\textbf{Thresholding module.}
We then assess thresholding methods at the frame level (0 = normal sample, 1 = anomaly), using: \textbf{TPR}, \textbf{TNR}, \textbf{Balanced Accuracy} $=(\text{TPR}+\text{TNR})/2$, \textbf{Weighted Accuracy} = $\beta \cdot \text{TNR}+(1-\beta)\cdot\text{TPR}$ with $\beta=\text{number of normal samples}/\text{total number of samples}$.

\textbf{End-to-end evaluation.}
Finally, we evaluate the full failure detection pipeline (including semantic filtering), where metrics only consider failure-related anomalies as positive samples. Metrics (TPR, TNR, Balanced/Weighted and Accuracy) are reported at observation level to assess the ability to detect true failures while ignoring benign anomalies.

% ============= Threshold results =============

\begin{figure*}[!t]
    \centering
    \includegraphics[width=\textwidth]{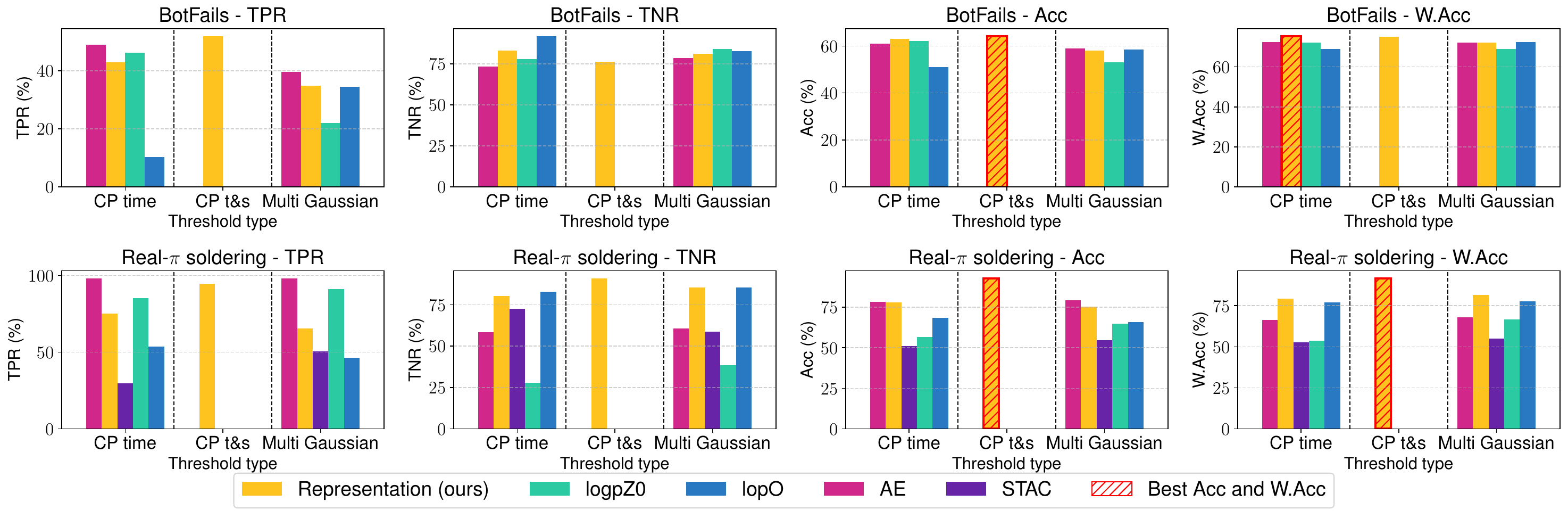}
    \caption{\textbf{Anomaly Detection evaluation with thresholding} - mean True Positive Rate (TPR), True Negative Rate (TNR), balanced accuracy (Acc), weighted accuracy (W.Acc), (\textbf{in \%}), for various Anomaly Detection types and various thresholding types--temporal Conformal Prediction (CP time), temporal and spatial Conformal Prediction (CP t\&s), Multiple Gaussian--evaluated on real world task from the BotFails dataset and soldering task operated autonomously with ACT \cite{zhao2023learningfinegrainedbimanualmanipulation}. \cellcolor{red!20}{Best Acc and W.Acc in hatched red}.}
    \label{fig:threshold_results}
\end{figure*}

% ============== Final Results ===============
\begin{figure*}[!t]
    \centering
    \includegraphics[width=\linewidth]{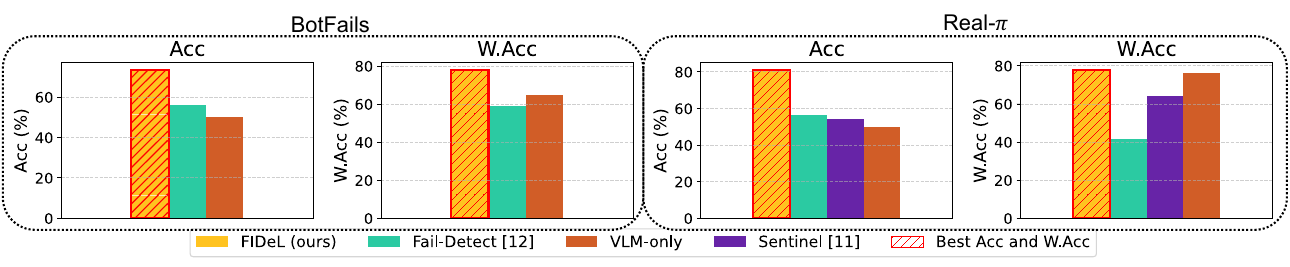}
    \caption{\textbf{End-to-end system evaluation (including semantic filtering)} - accuracy (Acc) and weighted accuracy (W.Acc) for various robotics failure detection systems. The labeling is different for this evaluation. While AD and Threshold module are evaluated using \textbf{anomaly labels}, the end-to-end system is evaluated using \textbf{failures labels}. Which means that all failure-related anomaly labels remain 1, while non-failure anomalies are relabeled from 1 to 0. FiDeL = Representation + CP t\&s + Semantic filtering
    }
    \label{fig:final_histo}
\end{figure*}

% =============================================

\subsection{Results and discussion}

Table~\ref{tab:AD_results} reports the performance of anomaly detection (AD) methods evaluated directly on raw anomaly scores. Our representation-based approach consistently achieves the best AUROC and F1@Opt across both datasets, surpassing all baselines by a clear margin. On BotFails, it yields a $+5.3\%$ AUROC gain over the second-best method (logpZ0), confirming robustness to the high semantic variability of anomalies. On the Real-$\pi$ soldering task, it further widens the gap, reaching $86.7\%$ AUROC versus $82.6\%$ for logpZ0 and $53.6\%$ for STAC. These results show that compact nominal representations outperform reconstruction- and likelihood-based methods in capturing deviations from expert behavior.

Figure~\ref{fig:threshold_results} compares thresholding strategies applied to anomaly scores. The best results are obtained with our representation-based method combined with CP-time\&space, which yields robust performance across datasets. Conformal prediction (CP) generally outperforms Gaussian thresholding, as it makes no distributional assumption and adapts to diverse score behaviors. In contrast, Gaussian fitting assumes Gaussian-distributed scores, an approximation that often fails under extreme conditions, leading to higher variance and degraded weighted accuracy—especially for logpZ0, where over-flagging anomalies reduces TNR. Still, Gaussian thresholding is not entirely ineffective: in some tasks where the score distribution is closer to normal, it can match or even surpass CP-time. Comparing CP variants, CP-time\&space consistently improves over CP-time. On Real-$\pi$, the gain is marked since anomalies in soldering are spatially localized. On BotFails, where anomalies are more diverse, the improvement is smaller: CP-time\&space still yields higher accuracy, though its weighted accuracy is marginally lower (–0.17\%), reflecting a TPR–TNR trade-off.

End-to-end results (Figure~\ref{fig:final_histo}) highlight the importance of semantic filtering when converting anomaly labels to failure labels. AD modules suffer a sharp performance drop when benign anomalies are relabeled as nominal, especially in TNR; for example, our Representation method with CP-time\&space loses $23\%$ TNR on Real-$\pi$. Adding semantic filtering significantly mitigates the TNR degradation, enabling more balanced detection ($85.8\%$ TPR / $74.8\%$ TNR). Fail-Detect tends to overpredict anomalies, as it cannot distinguish true failures from benign deviations, while Sentinel favors nominal predictions, likely missing failures that are purely visual and not reflected in proprioceptive signals. These results show that semantic filtering is crucial for reliable failure monitoring—avoiding unnecessary interventions while still capturing genuine failures.

%% file: sec/5_conclusion.tex
\section{CONCLUSIONS}

We introduced FIDeL, a representation-based anomaly detection module tailored for IL robotic policies. By combining optimal transport alignment, spatially-aware conformal thresholds, and semantic filtering, our approach enables reliable failure detection with interpretable localization. Experiments across diverse tasks show that FIDeL balances robustness to natural task variability with sensitivity to failure related deviations, highlighting the benefit of spatially-extended conformal prediction. Beyond detecting anomalies, FIDeL provides a practical interface for monitoring robot behavior in deployment, where reliable monitoring hinges not only on policy quality but also on the ability to recognize and interpret failures.

\section{Limitations}
\label{sec:limitations}

FIDeL shows strong failure detection in real-world robotic tasks, but several limitations remain. First, its performance depends heavily on the diversity of demonstrations, which may cause benign variations to be flagged as anomalies. While the VLM filter helps reduce false positives, its use is computationally costly, and excessive false alarms would slow inference. Second, the method’s spatial and temporal invariance, though useful for robustness, can reduce sensitivity to fine-grained or order-dependent deviations. This is particularly problematic for non-Markovian anomalies, where the correctness of an action depends not only on the current state but also on the history of states or actions that preceded it. (e.g., press buttons to perform a specific sequence). Detecting such cases requires explicit temporal modeling. Finally, balancing invariance and sensitivity remains task-dependent: some applications benefit from tolerance to small perturbations, while others demand detection of subtle deviations such as misalignments or object misplacements.

%% file: refs.bib
@article{xu2025FailDetect,
    title={Can We Detect Failures Without Failure Data? Uncertainty-Aware Runtime Failure Detection for Imitation Learning Policies},
    author={Xu, Chen and others},
    journal={arXiv preprint arXiv:2503.08558},
    year={2025}
  }

@inproceedings{
he2024rediffuser,
title={ReDiffuser: Reliable Decision-Making Using a Diffuser with Confidence Estimation},
author={Nantian He and others},
booktitle={Proc. of ICML},
year={2024},
}

@inproceedings{wang2024groundinglanguageplansdemonstrations,
      title={Grounding Language Plans in Demonstrations Through Counterfactual Perturbations}, 
      author={Yanwei Wang and Tsun-Hsuan Wang and Jiayuan Mao and Michael Hagenow and Julie Shah},
      year={2024},
      booktitle={Proc. of ICLR}, 
}

@inproceedings{agia2024STAC,
      title={Unpacking Failure Modes of Generative Policies: Runtime Monitoring of Consistency and Progress},
      author={Christopher Agia and others},
      year={2025},
      booktitle={Proc. of CoRL}, 
}

@inproceedings{
sun2023conformal,
    title={Conformal Prediction for Uncertainty-Aware Planning with Diffusion Dynamics Model},
    author={Jiankai Sun and others},
    booktitle={Proc. of NeurIPS},
    year={2023},
}

@inproceedings{guptaDetectingMitigatingSystemLevel2024,
  title = {Detecting and {{Mitigating System-Level Anomalies}} of {{Vision-Based Controllers}}},
  author = {Gupta, Aryaman and Chakraborty, Kaustav and Bansal, Somil},
  year = {2024},
  booktitle={Proc. of ICRA}
}

@misc{hafezSafeLLMControlledRobots2025,
  title = {Safe {{LLM-Controlled Robots}} with {{Formal Guarantees}} via {{Reachability Analysis}}},
  author = {Hafez, Ahmad and Akhormeh, Alireza Naderi and Hegazy, Amr and Alanwar, Amr},
  year = {2025},
  url={https://arxiv.org/abs/2503.03911}
}

@inproceedings{liu2023modelbasedruntimemonitoringinteractive,
      title={Model-Based Runtime Monitoring with Interactive Imitation Learning}, 
      author={Huihan Liu and Shivin Dass and Roberto Martín-Martín and Yuke Zhu},
      year={2024},
      booktitle={Proc. of ICRA}
}

@inproceedings{liuMultiTaskInteractiveRobot2024,
  title = {Multi-{{Task Interactive Robot Fleet Learning}} with {{Visual World Models}}},
  author = {Liu, Huihan and others},
  year = {2024},
  booktitle={Proc. of CoRL}

}

@misc{gokmen2023askinghelpfailureprediction,
      title={Asking for Help: Failure Prediction in Behavioral Cloning through Value Approximation}, 
      author={Cem Gokmen and Daniel Ho and Mohi Khansari},
      year={2023},
      url={https://arxiv.org/abs/2302.04334}, 
}

@inproceedings{chen2019neuralordinarydifferentialequations,
      title={Neural Ordinary Differential Equations}, 
      author={Ricky T. Q. Chen and Yulia Rubanova and Jesse Bettencourt and David Duvenaud},
      year={2018},
      booktitle={Proc. of NeurIPS} 
}

@misc{cadene2024lerobot,
    author = {Cadene, Remi and Alibert, Simon and Soare, Alexander and Gallouedec, Quentin and Zouitine, Adil and Wolf, Thomas},
    title = {LeRobot: State-of-the-art Machine Learning for Real-World Robotics in Pytorch},
    howpublished = "\url{https://github.com/huggingface/lerobot}",
    year = {2024}
}

@article{diffusionpolicy,
	author = {Cheng Chi and Zhenjia Xu and Siyuan Feng and Eric Cousineau and Yilun Du and Benjamin Burchfiel and Russ Tedrake and Shuran Song},
	title ={Diffusion Policy: Visuomotor Policy Learning via Action Diffusion},
	journal = {The International Journal of Robotics Research},
	year = {2024},
}

@misc{pizeropointcinq,
      title={$\pi_{0.5}$: a Vision-Language-Action Model with Open-World Generalization}, 
      author={Physical Intelligence and Kevin Black and others},
      year={2025},
      eprint={2504.16054},
      archivePrefix={arXiv},
      primaryClass={cs.LG},
      url={https://arxiv.org/abs/2504.16054}, 
}

@article{oquab2024dinov2learningrobustvisual,
      title={DINOv2: Learning Robust Visual Features without Supervision}, 
      author={Maxime Oquab and others},
      year={2024},
      journal={TMLR}, 
}

@article{Cui_2023,
   title={A Survey on Unsupervised Anomaly Detection Algorithms for Industrial Images},
     author={Cui, Yajie and Liu, Zhaoxiang and Lian, Shiguo},
   journal={IEEE Access},  
   year={2023}
}

@misc{chalapathy2019deeplearninganomalydetection,
      title={Deep Learning for Anomaly Detection: A Survey}, 
      author={Raghavendra Chalapathy and Sanjay Chawla},
      year={2019},
      url={https://arxiv.org/abs/1901.03407}, 
}

@article{tax1999support,
  title        = {Support vector domain description},
  author       = {Tax, David M.\,J. and Duin, Robert P.\,W.},
  journal      = {Pattern Recognition Letters},
  year         = {1999},
  publisher    = {Elsevier},
}

@inproceedings{li2021cutpaste,
      title={CutPaste: Self-Supervised Learning for Anomaly Detection and Localization}, 
      author={Chun-Liang Li and Kihyuk Sohn and Jinsung Yoon and Tomas Pfister},
      year={2021},
      booktitle={Proc. of CVPR}, 
}

@inproceedings{zavrtanik2021draem,
      title={DRAEM -- A discriminatively trained reconstruction embedding for surface anomaly detection}, 
      author={Vitjan Zavrtanik and Matej Kristan and Danijel Skočaj},
      year={2021},
      booktitle={Proc. of ICCV}, 
}

@inproceedings{schluter2022,
      title={Natural Synthetic Anomalies for Self-Supervised Anomaly Detection and Localization},
      author={Hannah M. Schlüter and Jeremy Tan and Benjamin Hou and Bernhard Kainz},
      year={2022},
      booktitle={Proc. of ICCV}, 
}

@inproceedings{venkataramanan2020VAEandGAN,
      title={Attention Guided Anomaly Localization in Images}, 
      author={Shashanka Venkataramanan and Kuan-Chuan Peng and Rajat Vikram Singh and Abhijit Mahalanobis},
      booktitle={Proc. of ECCV}, 
      year={2020}
}

@misc{liu2021AEandGAN,
      title={Unsupervised Two-Stage Anomaly Detection}, 
      author={Yunfei Liu and Chaoqun Zhuang and Feng Lu},
      year={2021},
      url={https://arxiv.org/abs/2103.11671}, 
}

@article{Shi_2021AE,
   title={Unsupervised anomaly segmentation via deep feature reconstruction},
   journal={Neurocomputing},
   author={Shi, Yong and Yang, Jie and Qi, Zhiquan},
   year={2021},
}

@inproceedings{wang2021studentteacherfeaturepyramidmatching,
      title={Student-Teacher Feature Pyramid Matching for Anomaly Detection}, 
      author={Guodong Wang and Shumin Han and Errui Ding and Di Huang},
      year={2021},
      booktitle={Proc. of BMVC}, 
}

@inproceedings{yamada2022studentteacherdiscriminativenetworks,
      title={Reconstructed Student-Teacher and Discriminative Networks for Anomaly Detection}, 
      author={Shinji Yamada and Satoshi Kamiya and Kazuhiro Hotta},
      year={2022},
      booktitle={Proc. of IROS}
}

@inproceedings{rudolph2022asymmetricstudentteachernetworksindustrial,
      title={Asymmetric Student-Teacher Networks for Industrial Anomaly Detection}, 
      author={Marco Rudolph and Tom Wehrbein and Bodo Rosenhahn and Bastian Wandt},
      year={2023},
      booktitle={Proc. of the IEEE/CVF winter conference on applications of computer vision}
}

@inproceedings{gudovskiy2021NF,
      title={CFLOW-AD: Real-Time Unsupervised Anomaly Detection with Localization via Conditional Normalizing Flows}, 
      author={Denis Gudovskiy and Shun Ishizaka and Kazuki Kozuka},
      year={2021},
     booktitle={Proc. of the IEEE/CVF winter conference on applications of computer vision}, 
}

@inproceedings{rudolph2021NF,
      title={Fully Convolutional Cross-Scale-Flows for Image-based Defect Detection}, 
      author={Marco Rudolph and Tom Wehrbein and Bodo Rosenhahn and Bastian Wandt},
      year={2021},
      booktitle={Proc. of the IEEE/CVF winter conference on applications of computer vision} 
}

@misc{yu2021fastflow,
      title={FastFlow: Unsupervised Anomaly Detection and Localization via 2D Normalizing Flows}, 
      author={Jiawei Yu and others},
      year={2021},
      url={https://arxiv.org/abs/2111.07677}, 
}

@inproceedings{defard2020padim,
      title={PaDiM: a Patch Distribution Modeling Framework for Anomaly Detection and Localization}, 
      author={Thomas Defard and Aleksandr Setkov and Angelique Loesch and Romaric Audigier},
      year={2020},
      booktitle={Proc. of ICPR}
}

@inproceedings{roth2022,
      title={Towards Total Recall in Industrial Anomaly Detection}, 
      author={Karsten Roth and others},
      year={2022},
      booktitle={Proc. of CVPR}
}

@inproceedings{Wang2021,
author = {Wang, Shenzhi and Wu, Liwei and Cui, Lei and Shen, Yujun},
year = {2021},
title = {Glancing at the Patch: Anomaly Localization with Global and Local Feature Comparison},
booktitle={Proc. of CVPR}
}

@inproceedings{zheng2022,
      title={Focus Your Distribution: Coarse-to-Fine Non-Contrastive Learning for Anomaly Detection and Localization}, 
      author={Ye Zheng and others},
      year={2022},
      booktitle={Proc. of ICME}
}

@article{abdalla2024videoanomalydetection10,
  author  = {Abdalla, Moshira and Javed, Sajid and Al Radi, Muaz and Ulhaq, Anwaar and Werghi, Naoufel},
  title   = {Video anomaly detection in 10 years: a survey and outlook},
  journal = {Neural Computing and Applications},
  year    = {2025},
  volume  = {37},
  number  = {32},
  pages   = {26321--26364},
  month   = nov,
  doi     = {10.1007/s00521-025-11659-8},
  url     = {https://doi.org/10.1007/s00521-025-11659-8},
  issn    = {1433-3058},
}

@misc{wu2024deeplearningvideoanomaly,
      title={Deep Learning for Video Anomaly Detection: A Review}, 
      author={Peng Wu and Chengyu Pan and Yuting Yan and Guansong Pang and Peng Wang and Yanning Zhang},
      year={2024},
      archivePrefix={arXiv},
      url={https://arxiv.org/abs/2409.05383}, 
}

@inproceedings{hasan2016learningtemporalregularityvideo,
      title={Learning Temporal Regularity in Video Sequences}, 
      author={Mahmudul Hasan and others},
      year={2016},
      booktitle={Proc. of CVPR}
}

@article{Wang2018,
author = {Wang, Tian and others},
year = {2018},
title = {Generative Neural Networks for Anomaly Detection in Crowded Scenes},
journal = {IEEE TIFS},
}

@article{fan2018videoanomalydetectionlocalization,
      title={Video Anomaly Detection and Localization via Gaussian Mixture Fully Convolutional Variational Autoencoder}, 
      author={Yaxiang Fan and others},
      year={2020},
      journal={Computer Vision and Image Understanding},
}

@inproceedings{rezende2016NF,
      title={Variational Inference with Normalizing Flows}, 
      author={Danilo Jimenez Rezende and Shakir Mohamed},
      year={2015},
      booktitle={Proc. of ICML}
}

@inproceedings{xu2024normalizingflowneuralnetworks,
      title={Normalizing flow neural networks by JKO scheme}, 
      author={Chen Xu and Xiuyuan Cheng and Yao Xie},
      year={2023},
      booktitle={Proc. of NeurIPS},
}

@inproceedings{he2015deepresiduallearningimage,
      title={Deep Residual Learning for Image Recognition}, 
      author={Kaiming He and Xiangyu Zhang and Shaoqing Ren and Jian Sun},
      year={2016},
      booktitle={Proc. of CVPR}
}

@inproceedings{Gal2016Dropout,
  title={Dropout as a Bayesian Approximation: Representing Model Uncertainty in Deep Learning},
  author={Gal, Yarin and others},
  booktitle={Proc. of ICML},
  year={2016}
}

@inproceedings{Kendall2017WhatUncertainties,
  title={What Uncertainties Do We Need in Bayesian Deep Learning for Computer Vision?},
  author={Kendall, Alex and Gal, Yarin},
  booktitle={Proc. of NeurIPS},
  year={2017}
}

@inproceedings{lockwood2022reviewuncertaintydeepreinforcement,
      title={A Review of Uncertainty for Deep Reinforcement Learning}, 
      author={Owen Lockwood and Mei Si},
      year={2022},
      booktitle={Proc. of AAAI}
}

@article{Zhu2019UQRL,
      title={Uncertainty Quantification and Exploration for Reinforcement Learning}, 
      author={YI Zhu and Jing Dong and Henry Lam},
      year={2024},
      journal={Operations Research}
}

@INPROCEEDINGS{zhao2023learningfinegrainedbimanualmanipulation, 
    AUTHOR    = {Tony Z. Zhao AND Vikash Kumar AND Sergey Levine AND Chelsea Finn}, 
    TITLE     = {{Learning Fine-Grained Bimanual Manipulation with Low-Cost Hardware}}, 
    BOOKTITLE = {Proceedings of Robotics: Science and Systems}, 
    YEAR      = {2023}, 
    ADDRESS   = {Daegu, Republic of Korea}, 
    MONTH     = {July}, 
    DOI       = {10.15607/RSS.2023.XIX.016} 
}

@misc{lei2017distributionfreepredictiveinferenceregression,
      title={Distribution-Free Predictive Inference For Regression}, 
      author={Jing Lei and Max G'Sell and Alessandro Rinaldo and Ryan J. Tibshirani and Larry Wasserman},
      year={2017},
      eprint={1604.04173},
      archivePrefix={arXiv},
      primaryClass={stat.ME},
      url={https://arxiv.org/abs/1604.04173}, 
}

@misc{diquigiovanni2021importancebandfinitesampleexact,
      title={The Importance of Being a Band: Finite-Sample Exact Distribution-Free Prediction Sets for Functional Data}, 
      author={Jacopo Diquigiovanni and Matteo Fontana and Simone Vantini},
      year={2021},
      eprint={2102.06746},
      archivePrefix={arXiv},
      primaryClass={stat.ME},
      url={https://arxiv.org/abs/2102.06746}, 
}

@misc{qwen2025qwen25technicalreport,
      title={Qwen2.5 Technical Report}, 
      author={Qwen and : and An Yang and Baosong Yang et al},
      year={2025},
      eprint={2412.15115},
      archivePrefix={arXiv},
      primaryClass={cs.CL},
      url={https://arxiv.org/abs/2412.15115}, 
}

@inproceedings{papagiannis2022imitationlearningsinkhorndistances,
      title={Imitation Learning with Sinkhorn Distances}, 
      author={Georgios Papagiannis and Yunpeng Li},
      booktitle={ECML/PKDD},
      year={2022},
      eprint={2008.09167},
      url={https://arxiv.org/abs/2008.09167}, 
}

@misc{dadashi2021primalwassersteinimitationlearning,
      title={Primal Wasserstein Imitation Learning}, 
      author={Robert Dadashi and Léonard Hussenot and Matthieu Geist and Olivier Pietquin},
      year={2021},
      eprint={2006.04678},
      archivePrefix={arXiv},
      primaryClass={cs.LG},
      url={https://arxiv.org/abs/2006.04678}, 
}

@INPROCEEDINGS{10208652,
  author={Chiu, Li-Ling and Lai, Shang-Hong},
  booktitle={2023 IEEE/CVF Conference on Computer Vision and Pattern Recognition Workshops (CVPRW)}, 
  title={Self-Supervised Normalizing Flows for Image Anomaly Detection and Localization}, 
  year={2023},
  volume={},
  number={},
  pages={2927-2936},
  doi={10.1109/CVPRW59228.2023.00294}}

@misc{bouman2025autoencodersanomalydetectionunreliable,
      title={Autoencoders for Anomaly Detection are Unreliable}, 
      author={Roel Bouman and Tom Heskes},
      year={2025},
      eprint={2501.13864},
      archivePrefix={arXiv},
      primaryClass={cs.LG},
      url={https://arxiv.org/abs/2501.13864}, 
}

@inproceedings{kirichenko2020flowood,
 author = {Kirichenko, Polina and Izmailov, Pavel and Wilson, Andrew G},
 booktitle = {Advances in Neural Information Processing Systems},
 pages = {20578--20589},
 publisher = {Curran Associates, Inc.},
 title = {Why Normalizing Flows Fail to Detect Out-of-Distribution Data},
 url = {https://proceedings.neurips.cc/paper_files/paper/2020/file/ecb9fe2fbb99c31f567e9823e884dbec-Paper.pdf},
 volume = {33},
 year = {2020}
}
